\documentclass[11pt]{article}
\usepackage{coling2020}
\usepackage{times}
\usepackage{url}
\usepackage{latexsym}
\usepackage[title]{appendix}

\colingfinalcopy

\title{Situated Data, Situated Systems: A Methodology to Engage with Power Relations in Natural Language Processing Research}

\author{Lucy Havens\textsuperscript{{\footnotesize$\dag$}}~~~~Melissa Terras \textsuperscript{{\footnotesize$\ddag$}}~~~~Benjamin Bach\textsuperscript{{\footnotesize$\dag$}}~~~~Beatrice Alex\textsuperscript{{\footnotesize$\S\dag$}}
\vspace{.1cm} \\
\textsuperscript{{\footnotesize$\dag$}}School of Informatics \\
\textsuperscript{{\footnotesize$\ddag$}}College of Arts, Humanities and Social Sciences \\
\textsuperscript{{\footnotesize$\S$}}Edinburgh Futures Institute; School of Literatures, Languages and Cultures \\
University of Edinburgh \\
\texttt{lucy.havens@ed.ac.uk, m.terras@ed.ac.uk} \\
\texttt{bbach@inf.ed.ac.uk, balex@ed.ac.uk} \\}

\date{November 3, 2020}

\begin{document}
\maketitle
\begin{abstract}
  We propose a bias-aware methodology to engage with power relations in natural language processing (NLP) research. NLP research rarely engages with bias in social contexts, limiting its ability to mitigate bias. While researchers have recommended actions, technical methods, and documentation practices, no methodology exists to integrate critical reflections on bias with technical NLP methods. In this paper, after an extensive and interdisciplinary literature review, we contribute a bias-aware methodology for NLP research. We also contribute a definition of biased text, a discussion of the implications of biased NLP systems, and a case study demonstrating how we are executing the bias-aware methodology in research on archival metadata descriptions.
\end{abstract}

\section{Introduction}

\blfootnote{\hspace{-0.65cm}This work is licensed under a Creative Commons Attribution 4.0 International License. License details: \url{http://creativecommons.org/licenses/by/4.0/}.}

Analysis of computer systems has raised awareness of their biases, prompting researchers to make recommendations to mitigate harms that biased computer systems cause. Analysis has shown computer systems exhibiting biases through racism\footnote{``A belief that one’s own racial or ethnic group is superior'' \cite{OED_racism}.} \cite{Noble_2018}, sexism\footnote{``[P]rejudice, stereotyping, or discrimination, typically against women, on the basis of sex'' \cite{OED_sexism}.} \cite{Perez_2019}, and classism\footnote{``The belief that people can be distinguished or characterized, esp.~as inferior, on the basis of their social class'' \cite{OED_classism}.} \cite{D_K_2020}. This list of harms is not exhaustive; biased computer systems may also harm people based on ability, citizenship, and any other identity characteristic. To mitigate harms from biased computer systems, researchers have recommended actions, methods, and practices. However, none of the recommendations comprehensively address the complexity of the problems bias causes.

Considering the numerous \emph{types} of bias that may enter a natural language processing (NLP) system, \emph{places} that bias may enter, and \emph{harms} that bias may cause, we propose a bias-aware methodology to comprehensively address the consequences of bias for NLP research. Our methodology integrates critical reflection on social influences on and implications of NLP research with technical NLP methods. To scope our research direction and inform our methodology, we draw on an interdisciplinary selection of literature that includes work from the humanities, arts, and social sciences. We intend the methodology to (a) support the reproducibility of NLP research, enabling researchers to better understand which perspectives were considered in the research; and (b) diversify perspectives in NLP systems, guiding researchers in explicitly communicating the social context their research so others can situate future research in contexts that have yet to be investigated.

We begin with our bias statement (§\ref{sec:bias}) and motivations for proposing a bias-aware NLP research methodology (§\ref{sec:why}). Next, we summarize the interdisciplinary literature informing our methodology (§\ref{sec:lit}), explain the methodology (§\ref{sec:method}), and demonstrate it with a case study of our ongoing research with bias in archival metadata descriptions (§\ref{sec:case}). We end with a summary and vision for future NLP research (§\ref{sec:concl}).

\section{Bias Statement}\label{sec:bias}
We situate this paper in the United Kingdom (UK) in the 21\textsuperscript{st} century, writing as authors who primarily work as academic researchers. We identify as three females and one male; and as American, German, and Scots. Together we have experience in natural language processing, human-computer interaction, data visualization, digital humanities, and digital cultural heritage. In this paper, we propose a bias-aware methodology for NLP researchers. We define \textbf{biased language} as \textit{written or spoken language that creates or reinforces inequitable power relations among people, harming certain people through simplified, dehumanizing, or judgmental words or phrases that restrict their identity; and privileging other people through words or phrases that favor their identity}. Biased language causes representational harms~\cite{Vainapel_2015,Sweeney_2013}, or the restriction of a person's identity through the use of hyperbolic or simplistic language~\cite{Blodgett_2020,Talbot_2003}. NLP systems built on biased language become biased computer systems, which ``\textit{systematically} and \textit{unfairly discriminate} against certain individuals or groups of individuals in favor of others'' \cite[p.~332]{Friedman_1996}. Representational harms may cause inequitable system performance for different groups of people, leading to allocative harms~\cite{Zhang_2020,Noble_2018}, or the denial of a resource or opportunity~\cite{Blodgett_2020}. The people who experience harms from biased NLP systems varies with the context in which people use the system and with the language source on which the system relies. Moreover, people may not be aware they are being harmed given the black-box nature of many systems~\cite{koene_2017}.  That being said, whether or not people realize they are being prejudiced against, the people harmed will be those excluded from the most powerful social group.

\section{Why does NLP need a Bias-Aware Methodology?}\label{sec:why}
Statistics report a homogeneity of perspectives among students in computer-related disciplines that do not reflect the diversity of people affected by computer systems, risking a homogeneity of perspectives in the technology workforce and the computer systems that workforce develops. For academic year 2018/19, statistics on students in the UK\footnote{Situating our research in the UK, we reference statistics from the UK's Higher Education Statistical Agency (HESA).} report that the dominant group of people studying computer-related subjects overwhelmingly are white males without a disability.\footnote{\url{www.hesa.ac.uk/news/16-01-2020/sb255-higher-education-student-statistics/subjects}.} Moreover, differences in total numbers of surveyed students across identity characteristics (e.g.~sex, ethnicity, disability) skew the statistics in favor of those reported as white, male, and without a disability. Lack of diverse perspectives among students in computer-related disciplines may limit the diversity of perspectives in the workforce, where the development of NLP and other computer systems occurs. As of 2019, the Wise Campaign reported that women comprise 24\% of the core-STEM workforce in the UK.\footnote{\url{http://www.wisecampaign.org.uk/statistics/2019-workforce-statistics-one-}\newline\url{million-women-in-stem-in-the-uk/}} Lack of diverse perspectives in the development of NLP and other computer systems risks technological decisions that exclude groups of people (``technical bias''), as well as applications of computer systems that oppress groups of people (``emergent bias'') \cite{Friedman_1996}.

That being said, even if student demographics in NLP and computer-related disciplines become more balanced, the data underlying NLP systems will still cause bias. Theories of discourse state that language (written or spoken) reflects and reinforces ``society, culture and power'' \cite[p.~45]{Bucholtz_2003}. In turn, NLP systems built on human language reflect and reinforce power relations in society, inheriting biases in language \cite{Caliskan_2017} such as stereotypical expectations of genders \cite{Haines_Deaux_Lofaro_2016} and ethnicities \cite{Garg_2018}. Drawing on feminist theory, we argue that all language is biased, because language records human interpretations that are situated in a specific time, place, and worldview \cite{Haraway_1988}. Consequently, all NLP systems are subject to biases originating in the social contexts in which the systems are built (``preexisting bias'') \cite{Friedman_1996}. Psychology research suggests that biased language causes representational harms: Vainapel et al. \shortcite{Vainapel_2015} studied how masculine-generic language (e.g. ``he'') versus gender-neutral language (e.g. ``he or she'') affected participants' responses to questionnaires. The authors report that women gave themselves lower scores on intrinsic goal orientation and task value in questionnaires using masculine-generic language in contrast to questionnaires using gender-neutral language.\footnote{The authors report that men showed no difference in their intrinsic goal orientation and task value scores with masculine-generic versus gender-neutral language in the questionnaires; impacts on people who do not identify as either a man or a woman are unknown as the study groups participants into these two gender categories \cite{Vainapel_2015}.} The study provides an example of how biased language may harm select groups of people, because the participants reported as women experienced a restriction of their identity, influencing their behavior to conform to stereotypes. 

Acknowledging the harms of biased language and biased NLP systems, researchers have proposed approaches mitigating bias, though no approach has fully removed bias from an NLP dataset or algorithm. To mitigate bias in datasets, Webster et al. \shortcite{Webster_2018} produced a dataset of gendered ambiguous pronouns (GAP) to provide an unbiased text source on which to train NLP algorithms. However, the GAP dataset reverses gender roles, assuming that gender is a binary rather than a spectrum.\footnote{See HCI Guidelines for Gender Equity and Inclusivity at \url{www.morgan-klaus.com/gender-guidelines.html}.} Any NLP system that uses the GAP dataset thus adopts its preexisting gender bias. Efforts to mitigate bias in algorithms are similarly limited, focusing on technical performance rather than performance in social contexts. Zhao et al. \shortcite{Zhao_2018_Debias} describe an approach to debias word embeddings, writing, ``Finally we show that given sufficiently strong alternative cues, systems can ignore their bias'' (p.~16). However, the paper does not explain the intended social context in which to apply the authors' approach, risking emergent bias.\footnote{While earlier paragraphs in the paper indicate a focus on gender bias and stereotypes related to professional occupations, the authors do not define \emph{bias} or \emph{gender bias}, nor do they identify the types of \emph{systems} to which they refer.} Additionally, Gonen and Goldberg \shortcite{Gonen_Goldberg_2019} demonstrate how this debiasing approach hides, rather than removes, bias. In our bias-aware methodology, we describe documentation and user research practices that facilitate transparent communication of biases that may be present in NLP systems, facilitating reflection on how to include more diverse perspectives and empower underrepresented people.

\section{Interdisciplinary Literature Review} \label{sec:lit}
To inform our proposed bias-aware NLP research methodology, we draw on an interdisciplinary corpus of literature from computer science, data science, the humanities, the arts, and the social sciences.

NLP and ML scholars have recommended actions to diversify perspectives in technological research, recognizing the value of diversity to bias mitigation. Blodgett et al. \shortcite{Blodgett_2020} and Crawford \shortcite{Crawford_2017} recommend interdisciplinary collaboration so researchers can learn from humanistic, artistic, and sociological disciplines regarding human behavior, helping researchers to more effectively anticipate harms that computer systems may cause, in addition to benefits they may bring, addressing risks of emergent bias. They also recommend engaging with the people affected by NLP and other computer systems, testing on more diverse populations to address the risk of technical bias, and rethinking power relations between those who create and those who are affected by computer systems to address the risk of preexisting bias. Though these recommendations address the three types of bias that may enter an NLP system, they do not articulate how to identify relevant people to include in the development and testing of NLP systems. Our bias-aware methodology builds on recommendations from Blodgett et al. \shortcite{Blodgett_2020} and Crawford \shortcite{Crawford_2017} by outlining how to identify and include stakeholders in NLP research (§\ref{sec:epr}).

D'Ignazio and Klein \shortcite{D_K_2020} propose data feminism as an approach to addressing bias in data science. They define data feminism as, ``a way of thinking about data, both their uses and their limits, that is informed by direct experience, by a commitment to action, and by intersectional feminist thought'' (p.~8).\footnote{Intersectionality refers to the way in which different combinations of identity characteristics from one individual to another result in different experiences of privilege and oppression \cite{Crenshaw_1991}. In feminist thought, multiple viewpoints are needed to understand reality; viewpoints that claim to be objective are, in fact, subjective, because knowledge is the result of human interpretation \cite{Haraway_1988}.} Data feminism has seven principles: examine power, challenge power, elevate emotion and embodiment, rethink binaries and hierarchies, embrace pluralism, consider context, and make labor visible. These principles facilitate critical reflection on the impacts of data's collection and use in social contexts. Our bias-aware methodology tailors these principles to NLP research, outlining activities that encourage researchers to consider influences on and implications of their work beyond the NLP community (§\ref{sec:epr}).

Within the NLP research community, Bender and Friedman \shortcite{Bender_2018} recommend improved documentation practices to mitigate emergent, technical, and preexisting biases. They recommend all NLP research includes a ``data statement,'' which they describe as, ``a characterization of a dataset that provides context to allow developers and users to better understand how experimental results might generalize, how software might be appropriately deployed, and what biases might be reflected in systems built on the software'' (p.~587). Aimed at developers and users of NLP systems, data statements reduce the risk of emergent bias. The authors also note: ``As systems are being built, data statements enable developers and researchers to make informed choices about training sets and to flag potential underrepresented populations who may be overlooked or treated unfairly'' (p.~599), helping authors of data statements reduce the risk of technical and preexisting biases. A data statement serves as guiding documentation for the case study approach we propose in our bias-aware methodology (§\ref{sec:ebf}), documenting the specific context in which NLP researchers work. Our bias-aware methodology guides research activities before, during, and after the writing of a data statement: for researchers reading data statements to find a dataset for an NLP system, our methodology guides their evaluation of a dataset's suitability for research; for researchers writing data statements, our methodology guides their documentation of the data collection process. 

In addition to technological disciplines, our methodology draws on critical discourse analysis~\cite{vanleeuwen_2009}, participatory action research~\cite{Reid_2008,Swantz_2008}, intersectionality~\cite{Crenshaw_1991,D_K_2020}, feminism~\cite{Haraway_1988,Harding_1995,Moore_2018}, and design~\cite{Martin_2012}. Participatory action research provides a way for NLP researchers to diversify perspectives in their research, engaging with the social context that influences and is affected by NLP systems. Intersectionality reminds researchers of the multitude of experiences of privilege and oppression that bias causes, because no single identity characteristic determines whether a person is ``dominant'' (favored) or ``minoritized'' (harmed) \cite{D_K_2020}. The case study approach common to design methods enables a researcher to make progress on addressing bias through explicitly situating research in a specific time and place, and conducting user research with people to understand their power relations in that time and place. Feminist theory values perspectives at the margins, encouraging researchers to engage with people who are excluded from the dominant group in a social context. Feminist theorist Harding~\shortcite{Harding_1995} writes, ``In order to gain a causal critical view of the interests and values that constitute the dominant conceptual projects...one must start from the lives excluded as origins of their design - from `marginal' lives'' (p.~341). Our bias-aware research methodology includes collaboration with people at the margins of NLP research in an effort to empower minoritized people.

\section{A Bias-aware Methodology}\label{sec:method}
 Our bias-aware methodology has three main activities: examining power relations (§\ref{sec:epr}), explaining the bias of focus (§\ref{sec:ebf}), and applying NLP methods (§\ref{sec:anm}). Though we discuss the activities individually, we recommend researchers execute them in parallel because each activity informs the others. We aim for the methodology to include activities that researchers may adapt to their own research context, be their focus on algorithm development, adaptation, or application; or on dataset creation. We hope for this paper to begin a dialogue on tailoring a bias-aware methodology to different types of NLP research.

\subsection{Examining Power Relations}\label{sec:epr}
\subsubsection*{Stakeholder Identification}\label{sec:si}
An NLP researcher executing the bias-aware methodology will document the distribution of power in the social context relevant to their research and language source. In the bias-aware methodology, a researcher considers language to be a partial record that provides knowledge situated in a specific time, place, and perspective. To understand which people's perspectives their language source (``the data'') includes and excludes, an NLP researcher will identify \textbf{stakeholders}, or those who are represented in, use, manage, or provide the data. Specifically, NLP research stakeholders are (1) the researcher(s), (2) producers of the data, (3) institutions providing access to the data, (4) people represented in the data, and (5) people who use the data. To investigate their stakeholders' power relations, an NLP researcher will observe who dominates the social setting(s) relevant to their research, and who experiences minoritization in the same setting(s). After identifying the stakeholders, the researcher will document their roles as dominant or minoritized, along with any limitations to their identification.

\subsubsection*{Stakeholder Collaboration}\label{sec:sc}
To understand how privilege and oppression are experienced among stakeholders, an NLP researcher will conduct \textbf{participatory action research} (PAR)~\cite{Reid_2008,Swantz_2008} with representative individuals from all five stakeholder groups. Researchers who conduct PAR attempt to establish collaborative relationships with representatives from their groups of stakeholders. Researchers are not experts bringing NLP systems to stakeholders; rather, researchers and stakeholders collaboratively study a social context to understand how NLP systems could empower people, particularly minoritized people. Instead of seeking an objective perspective, researchers foreground individual stakeholder perspectives, recording them as situated in a specific time and place, and using their multiplicity to gain insight into the complexity of the research's social context.  To understand how NLP research can empower people in a specific social context, we propose four \textbf{power relations questions} \footnote{We adapted these questions from Moore's work on feminist community archiving \cite{Moore_2018}.} for NLP researchers to answer: (1) who or what is included in the research, (2) who or what is excluded from the research, (3) how will the research define knowledge, and (4) who has agency and who can be empowered?

To understand the impacts of dominant people's interests and values, research following a bias-aware methodology will begin from the perspective of minoritized people, those who are typically excluded as a result (even if unintentional) of the interests and values of dominant people. The research will define knowledge as situated in specific times, places, and perspectives. The widespread availability of language as digital data may give the illusion of universal representation. However, critical discourse analysis reminds the NLP researcher that their data, composed of discourses,\footnote{``A connected series of utterances by which meaning is communicated'' \cite{OED_discourse}.} are ``socially constructed ways of knowing some aspect of reality'' \cite[p.~141]{vanleeuwen_2009}. Social hierarchies influence the data that becomes widely available, rendering minoritized groups of people invisible due to their exclusion from the data, or misrepresenting them due to their exclusion from the data collection process.

An NLP researcher will weigh insights gathered from different stakeholder groups equally, making the research's knowledge multi-faceted. Explicit documentation of the time, place, and perspective that produced the knowledge will inform future NLP research. Should a future researcher wish to reproduce the research, the documentation will guide the future researcher in seeking the proper social context. Should a future researcher wish to build upon the research, they will be able to compare and contrast the research's social setting with their own, guiding them in determining potential contributions. 

\subsubsection*{Unavailable Stakeholders}
In situations where the researcher cannot conduct PAR with stakeholders, the researcher will write a data biography.\footnote{We All Count has a free, interactive data biography tool at \url{wac-survey-rails.herokuapp.com}.} A data biography documents where data were collected and stored, who collected and owns the data, and why, when, and how the data were collected \cite{Krause_2019}. Writing a data biography facilitates critical reflection on the social influences on and social implications of a dataset, informing technical decisions when applying NLP methods. Datasets may circulate oppression of minoritized groups through inclusion and through omission. The key to recognizing who is dominant and minoritized is understanding that an individual may be both; power relations vary with the context of research.

\subsection{Explaining the Bias of Focus}\label{sec:ebf}
When explaining the type of bias on which NLP research focuses, a researcher will provide a definition and explain how this type of bias relates to other types of bias. For example, AllSides.com's ratings may guide the classification of political bias in news,\footnote{See the Media Bias Ratings at \url{www.allsides.com/media-bias/media-bias-ratings}.} Hanson et al.'s~\shortcite{hanson_2015} Accessible Writing Guide may inform research with stakeholders who include people with disabilities, and Hitti et al.~\shortcite{Hitti_2019} provide a model for how to clearly define and classify gender bias in collaboration with interdisciplinary experts. Table \ref{tax-table} provides examples of gender biased language organized into their gender bias taxonomy. When following the bias-aware methodology, NLP research to create annotated datasets for other types of bias will similarly include collaboration with relevant disciplinary experts (i.e.~racial bias with critical race theory experts) to define and categorize types of bias relevant to the research. When writing a data statement's \textit{curation rationale}, an NLP researcher will include a definition of their bias of focus. In the answers to the power relations questions, an NLP researcher will describe how they consider intragroup differences within their stakeholder groups, in addition to differences between dominating and minoritized stakeholder groups, because the intersection of identity characteristics, rather than one identity characteristic in isolation, determines how people experience oppression \cite{Crenshaw_1991}. Due to the complexity that intersecting identity characteristics add to evaluations of bias, in the bias-aware methodology, an NLP researcher will use case studies.
\begin{table*}
\centering
\begin{tabular}{p{3.4cm}p{3.4cm}}
\hline
\multicolumn{2}{c}{\textbf{Structural Bias}}\\
\hline 
Gender \newline Generalization & \emph{\textbf{A lawyer} must always carry \textbf{his} phone.}\newline\newline\\
Explicit Marking \newline of Sex & \emph{The role of \textbf{a waitress} is overlooked by the restaurant owners.}\\
\hline
\end{tabular}
\begin{tabular}{p{3.4cm}p{3.4cm}}
\hline
\multicolumn{2}{c}{\textbf{Contextual Bias}}\\
\hline 
Societal Stereotype & \emph{The event was \textbf{sports-themed} for all the\newline \textbf{fathers} volunteering.}\newline\\
Behavioral\newline Stereotype & \emph{\textbf{All girls} are \textbf{sensitive}.}\newline\newline\\
\hline
\end{tabular}
\caption{\label{tax-table} Biased text examples classified into the gender bias taxonomy of Hitti et al.~\shortcite{Hitti_2019}.}
\end{table*}
Case studies gather information in a clearly-defined context and present the resulting knowledge as connected to a specific time, place, and people. To conduct a case study, an NLP researcher will ``determine a problem, make initial hypotheses, conduct research through interviews, observations, and other forms of information gathering [such as PAR], revise hypotheses and theory, and tell a story'' \cite[p.~28]{Martin_2012}. Feminist theory's focus on agency and lived experience as situated in a specific context adds value to PAR by helping a researcher anticipate and critically examine the implications of PAR's drive towards action \cite{Reid_2008}. When documenting their case study in blogs, presentations, or publications, an NLP researcher will discuss potential applications of the research beyond the case study's context, anticipating potential benefits and harms. Potential harms may outweigh potential benefits, making the best decision not to build an NLP system \cite{Crawford_2017}.

\subsection{Applying NLP Methods}\label{sec:anm}
When applying NLP methods in the bias-aware methodology, an NLP researcher should acknowledge biases found with any algorithms they use in their data statement. For example, when applying word embeddings, an NLP researcher could look to Bolukbasi et al.~\shortcite{Bolukbasi_2016}, Caliskan et al.~\shortcite{Caliskan_2017}, and Kurita et al.~\shortcite{Kurita_2019} on gender bias; Swinger et al.~\shortcite{Swinger_2019} on racial bias; Diaz et al.~\shortcite{Diaz_2018} on age bias; Papakyriakopoulos~\shortcite{Papakyriakopoulos_2020} on sexuality and nationality bias; and Gonen and Goldberg~\shortcite{Gonen_Goldberg_2019} on the inadequacy of debiasing word embeddings. When applying part-of-speech tagging, dependency parsing, or machine translation, an NLP researcher could look to Garimella et al.~\shortcite{Garimella_2019} and Stanovsky et al.~\shortcite{Stanovsky_2019} for understanding how these methods have been shown to exhibit gender bias. If an NLP researcher will train an algorithm on their language source, research documentation will describe the training process and results. If the research includes annotation, documentation will include instructions given to annotators.

For NLP research on algorithms, we recommend considering approaches to making bias transparent, in addition to reducing the biased behavior of algorithms. Research from Kaneko et al.~\shortcite{Kaneko_2019} and Zhao et al.~\shortcite{Zhao_2018_Debias} on mitigating bias in word embeddings provide starting points for algorithmic bias research, as their methods have yet to be evaluated in diverse contexts. However, Gonen and Goldberg~\shortcite{Gonen_Goldberg_2019} have shown the limits of debiasing word embeddings. We argue that the situated nature of data, and thus the situated nature of knowledge drawn from data, makes the elimination of bias impossible. Investigating how to make bias transparent provides an alternative direction for NLP researchers interested in mitigating bias in NLP systems. Whether making bias transparent or reducing biased behavior of algorithms, NLP researchers following the bias-aware methodology will collaborate with relevant disciplinary experts and minoritized stakeholders in determining how to evaluate an algorithm for bias.

To support the training of algorithms in diverse contexts, NLP research on datasets will define the context of its language source's collection and annotation. An NLP researcher will provide data statements to inform algorithms' training and evaluation, ensuring reproducibility and avoiding unintended harms from misapplications of algorithms \cite{Bender_2018}. Similarly, dataset research will include disciplinary experts and minoritized stakeholders in datasets' creation, annotation, and evaluation.

\section{Case Study}\label{sec:case}
In this section we describe how we are implementing the bias-aware NLP research methodology in a case study on bias in metadata descriptions from the online archival catalog of the Centre for Research Collections at the University of Edinburgh (``the Archive'').\footnote{Metadata documents information about collections of cultural heritage records. Archival catalogs have numerous metadata fields that contain descriptions written by people who archives hire to document their collection items. These descriptions are the language source we refer to as \textit{archival metadata descriptions} \cite{Angel_2019}.} For consistency with the outline of a bias-aware methodology (§\ref{sec:method}), we group our case study into the same three activities, explaining our examination of power relations (§\ref{sec:case-epr}), our bias of focus (§\ref{sec:case-ebf}), and then our application of NLP methods (§\ref{sec:case-anm}). Each subsection includes accomplished, ongoing, and planned future work. To demonstrate how we execute the three activities in parallel, as proposed in §\ref{sec:method}, we first provide a chronological overview.

Initially, our research began with information gathering linked to a participatory action research (PAR) methodology. We reviewed literature on bias in NLP and archives, and on digital humanities research (collaborations between technologists and humanists that often analyze data sources with historical language). We also met with employees at the Archive to better understand the Archive's policies, which guide the writing of metadata descriptions and documentation practices, such as the metadata standards used. The employees described how they are proactively challenging the inherited metadata and inherited practices of the Archive, which date back to the 16\textsuperscript{th} century. After the literature review and meeting we began writing data statements for the Archive's metadata descriptions and for our research. Due to the limited research on NLP methods applied to archival metadata, and limited large-scale analysis of metadata descriptions, we undertook a pilot data project,\footnote{View the pilot in a Jupyter Notebook at \url{github.com/thegoose20/eula41}.} walking through the process of extracting metadata descriptions from a single archival collection, adding historical context to our documentation of the extracted descriptions, and calculating corpus analytics (using ElementTree\footnote{\url{docs.python.org/3/library/xml.etree.elementtree.html}} and NLTK\footnote{\url{www.nltk.org}} in a Jupyter Notebook\footnote{\url{jupyter.org}}). After establishing a workflow to extract metadata descriptions from the Archive's online catalog, we again met employees at the Archive to discuss the challenges that biased language poses to their work and to their visitors. This meeting helped us add to our data statements, identify stakeholders in our research, and begin describing the stakeholders' power relations. Moreover, the meeting confirmed the value of an NLP system that detects and classifies bias, as the Archive does not currently have a systematic approach to measuring bias in its catalog's metadata descriptions.

\subsection{Researcher and Archive Power Relations}\label{sec:case-epr}
\subsubsection*{Stakeholder Identification}
In our execution of the bias-aware methodology, we study power relations among five stakeholders: (1) us (the authors) as researchers, (2) the Archive's employees, (3) the Archive (as an institution), (4) people represented in metadata descriptions, and (5) the Archive's visitors. Literature on power relations in archives and the wider gallery, library, archive, and museum (GLAM) sector~\cite{Adler_2017,Caswell_2019,Hauswedell_2020,McPherson_2012,Risam_2015} informed our identification of these stakeholders. We recorded our understanding of their power relations in our data statement (Appendix \ref{sec:appendixa}) and power relations document (Appendix \ref{sec:appendixc}), and will continue expanding and revising these documents until our research ends.

\subsubsection*{Stakeholder Collaboration}
In line with PAR, we collaborate with stakeholders at the Archive to learn about their perception of biased language in metadata descriptions, as well as challenges and potential approaches to addressing the bias. Thus far, we facilitated a group discussion with stakeholders who had a range of roles, including technical, curatorial, administrative, servicing, and documenting responsibilities; and a range of GLAM work experience, from one year to over 20 years. The group discussion informs our understanding of the range of attitudes towards bias and neutrality in archival documentation. We are preparing a survey to study how the Archive's attitudes about bias and neutrality relate to those of other UK archives. Results of the group discussion enabled us to draft answers to the power relations questions. 

\subsubsection*{Unavailable Stakeholders}
To fully answer the power relations questions, we are researching historical changes in the structure of metadata standards used at the Archive. Our stakeholders include people who documented the Archive's collections but no longer work there, and people who are written about in the Archive's metadata, which document material dating back to the 1\textsuperscript{st} century AD. To study power relations among these unavailable stakeholders, we are writing a data biography (Appendix \ref{sec:appendixb}) for the metadata descriptions with the Archive. The data biography informs our understanding of the power relations at play in our research, which in turn informs our data statement and technical decisions about NLP methods to apply.

\subsection{Contextual Gender Bias as a Focus}\label{sec:case-ebf}
Our NLP research focuses on identifying types of contextual gender bias from archival metadata descriptions, complementing Hitti et al.'s~\shortcite{Hitti_2019} focus on identifying structural gender bias. We adopt the their taxonomy of gender bias (illustrated in Table \ref{tax-table}). The taxonomy has two subtypes of contextual bias: behavioral stereotypes and societal stereotypes. We may expand on definitions and subtypes of contextual bias during our research into simplistic, hyperbolic language in metadata descriptions that indicates the presence of stereotypes, because historical text often contains spellings and syntax (among other linguistic characteristics) different to the modern text on which NLP tools have been developed \cite{Casey_2020}. In the context of the Archive, gender biased metadata descriptions may cause representational harms, because the Archive supports information access, circulating ideas documented in its metadata when users search its online catalog. Societal and behavioral stereotypes present in the Archive's metadata descriptions may negatively impact perceptions of people represented in the descriptions. We are researching the types of gender bias in the descriptions, and ways to measure such biases, in an effort to support the Archive in mitigating harms from biased metadata descriptions.

\subsection{Information Extraction for Classification}\label{sec:case-anm}
\subsubsection*{Information Extraction Methods}
The archival metadata descriptions we use as this case study's language source are from the Archive's public, online catalog. We obtained descriptive metadata fields as Extensible Markup Language (XML) data using the Open Archives Initiative - Protocol for Metadata Harvesting (OAI-PMH),\footnote{\url{www.openarchives.org/OAI/2.0/openarchivesprotocol.htm}} filtered the metadata for descriptive fields relevant to our research, and then removed duplicate descriptions. Table \ref{data-table} summarizes the resulting corpus. The Archive organizes metadata hierarchically, creating metadata for collections, subcollections, and items; we group subcollection and item descriptions within their overarching collection. Currently, we are exploring how to further filter our extracted descriptions through a combination of historical research on archival metadata standards and corpus analytics of terms surrounding gender-related words (as in the third use case from Casey et al.~\shortcite{Casey_2020}). For example, the Archive uses Library of Congress Subject Headings (LCSH), which use terms offensive to certain social groups: Adler~\shortcite{Adler_2017} discusses how LCSH represents people who do not identify with binary genders or do not conform to heterosexuality as ``deviations.'' To further filter our extracted metadata descriptions, we can associate the descriptions with the dates they were written and look for offensive terms that were used in metadata standards during those dates. Our data statement further details this process.
\begin{table*}
\centering
\begin{tabular}{p{2.2cm}p{2.8cm}p{2.8cm}p{2.8cm}p{2.9cm}}
\hline
\textbf{By Metadata\newline Field} & Biographical/\newline Historical & Scope and\newline Contents & Processing\newline Information & Total (sum of the metadata fields)\\
\hline 
Sentences & 11,323 & 55,434 & 1,691 & 68,448\\
Words & 801,893 & 208,190 & 11,016 & 966,763\\
\end{tabular}
\end{table*}

\begin{table*}
\centering
\begin{tabular}{p{2.2cm}p{2.8cm}p{2.8cm}p{2.8cm}p{2.9cm}}
\hline
\textbf{By Collection} & Minimum & Maximum & Mean & Standard\newline Deviation\\
\hline
Words & 7 & 156,747 & 1,036.2 & 7,784.5\\
\end{tabular}
\caption{Words and sentences in the extracted metadata descriptions from the Archive's 1,231 collections, calculated using Punkt tokenizers in the Natural Language Toolkit Python library \cite{Bird_2002}.}\label{data-table}
\end{table*}

\subsubsection*{Annotations to Inform Classification}
With our case study, we aim to create and annotate a gold standard dataset on which we will train a classification algorithm to identify types of gender bias in text. We will perform the annotations as part of the research for a Doctor of Philosophy project. Due to ethical concerns regarding the use of crowdsourcing platforms \cite{Gleibs_2017}, anyone employed to contribute to the annotation work will be paid at least minimum wage. To guide the annotation process and ensure the reproducibility of our research, we will document instructions we follow to annotate contextual gender bias. We will collaborate with the Archive and a gender studies expert to write these instructions; we are in the process of finding a language expert with whom to collaborate. When we publish the results of our research, we will provide documentation of the annotation instructions, data statements, data biography, and power relations questions for our NLP research. After creating a gold standard dataset annotated for contextual gender bias, we plan to train a discriminative classifier on the dataset using supervised learning. We will then experiment with and evaluate how the classifier differentiates between types of contextual gender bias in archival metadata descriptions, and report openly on the results of this research.

\section{Conclusion}\label{sec:concl}
In this paper we propose a bias-aware methodology for NLP research to mitigate harms from biased NLP systems. The methodology integrates practices and methods from NLP, ML, data science, gender and feminist studies, linguistics, and design. Due to the numerous types of bias, the intersectional nature of oppression, and the possibility of direct and indirect harms from bias, detecting and measuring bias is a complex process. Our methodology encourages NLP researchers to situate their work in case studies, explicitly describing the context of and stakeholders in their research. We advise NLP researchers to build the time and resources needed to undertake such work into project plans, and to put eliminating bias at the center of their research. Documenting instances of bias and their associated power relations will enable the NLP community to look for patterns across different contexts that use NLP systems. Amassing case studies in order to look for such patterns will guide NLP research towards generalizable approaches to bias mitigation, approaches that do not unintentionally minoritize people whose perspectives were unknowingly excluded.

\section*{Acknowledgments}
This paper describes work conducted in collaboration with Rachel Hosker and her team at the Centre for Research Collections (CRC) at the University of Edinburgh. Hosker and her team are activists seeking to change archives' descriptive language and practices to more accurately and inclusively represent the diverse populations for whom their collections are intended. Before we joined them as collaborators, they were discussing and making changes to the Archive's descriptive language and practices. We are grateful for the willingness of Hosker and her team at the CRC to collaborate with us, bringing together the knowledge and practices of the archival and NLP communities to mitigate harms from biased language.

\begin{appendices}

    \section{Data Statement for Metadata Descriptions Extracted from the Archive's Online Catalog (version 1)}
    \label{sec:appendixa}
    \subsection{Curation Rationale}
    
    We (the research team) will use the extracted metadata descriptions to create a gold standard dataset annotated for contextual gender bias. We adopt Hitti et al.'s definition of contextual gender bias in text: written language that connotes or implies an inclination or prejudice against a gender through the use of gender-marked keywords and their context~\shortcite[p.~10-11]{Hitti_2019}.
    
    A member of our research team has extracted text from three descriptive metadata fields for all collections, subcollections, and items in the Archive’s online catalog. One of these fields provide information about the people, time period, and places associated with the collection, subcollection, or item to which the field belongs. Another field summarizes the contents of the collection, subcollection, or item to which the field belongs. The last field records the person who wrote the text for the collection, subcollection, or item’s descriptive metadata fields, and the date the person wrote the text.
    
    Using the dataset of extracted text, we will experiment with training a discriminative classification algorithm to identify types of contextual gender bias. Additionally, the dataset will serve as a source of annotated, historical text to complement datasets composed of contemporary texts (i.e. from social media, Wikipedia, news articles).\newline\newline
    \textit{To Do: We will group the metadata descriptions based on the collection to which they’re associated, rather than segmenting by sentence or paragraph for annotation. Prior to making annotations for contextual gender bias, a member of our research team will review a subset of the metadata descriptions to determine whether all the descriptions should be annotated or whether the dataset should be filtered to include only a portion of the extracted metadata descriptions. Section B. in our data biography describes our plans for filtering.}\newline
    
    \noindent We chose to use archival metadata descriptions as a data source because:
    \begin{enumerate}
      \item Metadata descriptions in the Archive’s catalog (and most GLAM catalogs) are freely, publicly available online
      \item GLAM metadata descriptions have yet to be analyzed at large scale using natural language processing (NLP) methods and, as records of cultural heritage, the descriptions have the potential to provide historical insights on changes in language and society \cite{Welsh_2016}
      \item GLAM metadata standards are freely, publicly available, often online, meaning we can use historical changes in metadata standards used in the Archive to guide large-scale text analysis of changes in the language of the metadata descriptions over time
      \item The Archive’s policy acknowledges its responsibility to address legacy descriptions in its catalogs that use language considered biased or otherwise inappropriate today\footnote{The Archive is not alone; across the GLAM sector, institutions acknowledge and are exploring ways to address legacy language in their catalogs’ descriptions. The ``Note'' in We Are What We Steal provides one example: \url{https://dxlab.sl.nsw.gov.au/we-are-what-we-steal/notes/}.}
    \end{enumerate}
    
    \subsection{Language Variety}
    
    The metadata descriptions extracted from the Archive’s catalog are written in British English.
    
    \subsection{Producer Demographic}
    
    We (the research team) are of American, German, and Scots nationalities, and are three females and one male. We all work primarily as academic researhers in the disciplines of natural language processing, data science, data visualization, human-computer interaction, digital humanities, and digital cultural heritage. Additionally, one of us is auditing an online course on feminist and social justice studies.
    
    \subsection{Annotator Demographic}
    
    For the research team who will write the annotation rule book, please refer to the previous section.
    
    A gender, sexuality, and social justice studies expert based at a North American university will collaborate with us (the research team) on writing the annotation rule book.  One member of our research team will annotate the metadata in collaboration with a second annotator.\newline\newline
    \textit{Ongoing: we are seeking a second annotator with a background in gender studies, linguistics, or the information sciences; or with GLAM work experience.}
    
    \subsection{Speech or Publication Situation}
    
    The metadata descriptions extracted from the Archive’s online catalog using Open Access Initiative - Protocol for Metadata Harvesting (OAI-PMH). For OAI-PMH, an institution (in this case, the Archive) provides a URL to its catalog that displays its catalog metadata in XML format.  A member of our research team wrote scripts in Python to extract three descriptive metadata fields for every collection, subcollection, and item in the Archive’s online catalog (the metadata is organized hierarchically). Using Python and its Natural Language Toolkit (NLTK) library, the researcher removed duplicate sentences and calculated that the extracted metadata descriptions consist of a total of 966,763 words and 68,448 sentences across 1,231 collections. The minimum number of words in a collection is 7 and the maximum, 156,747, with an average of 1,306 words per collection and standard deviation of 7,784 words.
    
    Please refer to the Provenance Appendix for information on the Speech or Publication Situation of all of the Archive’s metadata descriptions.
    
    \subsection{Data Characteristics}
    
    Upon extracting the metadata descriptions using OAI-PMH, the XML tags were removed so that the total words and sentences of the metadata descriptions could be calculated to ensure the text source provided a sufficiently large dataset. A member of our research team has grouped all the extracted metadata descriptions by their collection (the ``fonds'' level in the XML data), preserving the context in which the metadata descriptions were written and will be read by visitors to the Archive’s online catalog.
    
    \subsection{Data Quality}
    
    As a member of our research team extracts and filters metadata descriptions from the Archive’s online catalog, they write assertions and tests to ensure as best as possible that metadata isn’t being lost or unintentionally changed.
    
    Please refer to the Provenance Appendix for information on the Data Quality of all of the Archive’s metadata descriptions.
    
    \subsection{Other}
    
    Not applicable
    
    \subsection{Provenance Appendix}
    
    Data Statement for Metadata Descriptions from the Archive’s Online Catalog
    (version 1)
    
    \subsubsection*{Curation Rationale}
    
    The Archive’s policy describes a commitment to develop collections that are as inclusive and diverse as possible, keeping up with social changes and looking for opportunities to better represent communities of people. Additionally, the Archive’s policy states that the Archive aims to make its collections accessible to as many people as possible.\newline\newline
    \textit{To Do: If available, review historical policy documents to understand how the Archive’s curation rationale has evolved since its founding.}
    
    \subsubsection*{Language Variety}
    
    The Archive’s metadata descriptions are written in British English.
    
    \subsubsection*{Producer Demographic}
    
    People who write metadata descriptions to document the Archive’s collections include employees, interns, and volunteers. Employees have received professional training in archival documentation, in addition to training at the Archive. Interns and volunteers are typically students studying information sciences, museology, history, or related disciplines who have also received training at the Archive. The Archive began in the 16\textsuperscript{th} century, so the metadata descriptions in its online catalog date from that time period up through the present day (the Archive continues to collect and document cultural heritage records).
    
    Additional demographic information on all those who have written the Archive’s metadata descriptions is limited, however the Archive is based in the United Kingdom, meaning the perspectives of those who wrote the descriptions is most likely English, Irish, Scottish, British, or European. The Archive is closely associated with a research university, so interns and volunteers who write the Archive’s metadata descriptions are likely to have received, or be in the process of receiving, higher education degrees. 
    
    \subsubsection*{Annotator Demographic}
    
    Not applicable
    
    \subsubsection*{Speech or Publication Situation}
    
    The metadata descriptions in the Archive’s online catalog document collections created by a university associated with the Archive and acquired or donated from other people and organizations. The Archive’s earliest metadata descriptions were written in the 16\textsuperscript{th} century; metadata descriptions continue to be written today.
    
    The goal of the metadata descriptions is to help people find primary source material in the Archives. At the time most of the Archive’s metadata descriptions were written, the descriptions were intended for employees of the Archive, who would help visitors locate primary source material. Circa 2015, employees of the Archive began writing metadata descriptions with visitors included in their intended audience.
    
    Current employees at the Archives have stated that they would be happy for the metadata descriptions they write to be viewed as works in progress, because the Archive could never have enough time to document all its collection items completely. Moreover, often information about collections items is impossible to know due to their historical nature and lack of accompanying documentation, so the metadata descriptions will always be incomplete.
    
    The metadata descriptions include information available from the cultural heritage records they describe, from any available documentation that accompanied those records when the Archive acquired them, from authorities such as the Library of Congress Subject Headings, and from other documentation resources considered trustworthy among archives (a more extensive list is provided here).
    
    \subsubsection*{Data Characteristics}
    
    Beginning circa 2017, people documenting collections in the Archive have written metadata descriptions according to the General International Standard Archival Description (ISAD(G)). Past metadata descriptions were written according to library metadata standards. Metadata descriptions may include contextual information about the people, places, and time periods relevant to the collection items, as well as the date a description was written and who wrote the description. Though all of this descriptive information ideally exists for a collection item, some collection items do not have this complete of a description.\newline\newline
    \textit{To Do:~If possible, determine which library metadata standards were used for documentation prior to 2017.}
    
    \subsubsection*{Data Quality}
    
    The metadata descriptions in the Archive’s online catalog consists of manually entered data, some of which was initially written in digital form, and some of which was initially written on paper and has since been manually typed into digital form.\newline\newline
    \textit{To Do:~Determine how much the metadata descriptions are born-digital versus re-written digitally, and when the Archive transitioned from writing metadata descriptions on paper to writing metadata descriptions digitally (typing manually).}
    
    \subsubsection*{Other}
    
    Not applicable
    
    \subsubsection*{Provenance Appendix}
    
    None 
    
    \section{Data Biography for Metadata Descriptions Extracted from the Archive's Online Catalog (version 1)}
    \label{sec:appendixb}
    \subsection{Dataset}
    
    Metadata descriptions from the Archive’s online catalog
    
    \subsection{Where was the Data Collected or Created?}
    
    We (the research team) collected the data using the Open Access Initiative - Protocol for Metadata Harvesting (OAI-PMH).
    
    Employees, interns, and volunteers at the Archive who wrote the metadata descriptions collected information to include in the descriptions from documentation accompanying the cultural heritage record(s) they were describing, from the cultural heritage records themselves, from authorities such as Library of Congress Subject Headings, and from other trusted sources for archival documentation. Examples of other trusted sources are available here.
    
    Where possible, we will use dates associated with the descriptions to contextualize their text in relation to historical changes in metadata structures. For example, the metadata standard Library of Congress Subject Headings (LCSH) once used the term “Jewish Question” instead of the current term “Jews,” so GLAM who use LCSH may have descriptions in their catalogs that use the historical term now considered biased. After historical analysis of metadata standards the Archive uses, we will filter our collected text to include those that reference groups of people who have historically been described stereotypically.
    
    \subsection{Who Collected or Created the Data?}
    
    The Archive and the university to which it is associated collected some of the cultural heritage records and the accompanying documentation that informs the records’ metadata descriptions. For other cultural heritage records and their accompanying documentation, individual collectors gathered the records and wrote their documentation, which employees, interns, and volunteers used to write descriptive metadata for the records in the Archive’s catalog.
    
    The Archive has existed since the 16\textsuperscript{th} century, so its directors will each have established different policies and goals for acquiring and documenting cultural heritage records. The latest policy document for the Archive includes a statement about diversity, inclusion and accessibility that describes the Archive’s commitment to providing representative collections for local, national, and international audiences.
    
    \subsection{Why was the Data Collected or Created?}
    
    The Archive’s policy explains that it documents cultural heritage records in its catalog so that researchers can find the records and use them as primary source material to guide their work. Current employees of the Archive reiterated the goal of discoverability as the main reason for writing metadata descriptions.
    
    Individuals and institutions who have donated their collections to the Archive had personal reasons motivating their choices of records to save. A directory of the Archive’s collections contains information about select individuals and institutions that suggest their reasons for saving the records they did. Information in the metadata descriptions themselves may also provide insight on why their associated records were collected.
    
    \subsection{When was the Data Collected or Created?}
    
    Among the metadata descriptions we extracted that include a year documenting when they were written, the years show that the descriptions were written from the 19\textsuperscript{th} century up through the 21\textsuperscript{st} century. Further research is needed to determine how early the extracted metadata descriptions without a year were written.
    
    \section{Stakeholder Power Relations in NLP Research on Bias in Archival Metadata Descriptions (version 1)}
    \label{sec:appendixc}
    \subsection{The Stakeholders}
    
    \textbf{Identification:}
    \begin{enumerate}
      \item Us as the research team
      \item Employees of the Archive (current and former) who wrote the metadata descriptions that serve as this research’s text source
      \item The Archive and its associated university as institutions that provide access to the metadata descriptions
      \item People represented in the metadata descriptions
      \item Visitors to the Archive, as they will read the metadata descriptions used as this research’s text source when using the Archive’s online catalog
    \end{enumerate}
    \textbf{Limitations:}
    Due to the length of the text and the historical nature of the metadata descriptions we use from the Archive’s catalog, we do not have access to every person represented in the metadata descriptions. However, the Archive does have a take-down policy that we will follow with our text source to respect the people represented in metadata descriptions as best as possible: if a person requests that information about them or someone they are connected to be removed from or anonymized in the catalog, the Archive will comply. To the best of our ability, we will make sure that the metadata descriptions we use as the text source for our research do not include information that a visitor has requested the Archive take down.
    
    \subsection{Power Relations Questions}
    
    \subsubsection*{Who or what is included in the research?}
    
    Who:
    \begin{itemize}
      \item Current employees of the Archive: To account for intragroup differences, we include employees with different years of experience and employees working in several positions within the hierarchy of job roles in the Archive.
      \item Us (the research team): The size of the team is small enough that all members are included, meaning intragroup differences are accounted for by default.
    \end{itemize}
    \textit{To Do: Find visitors to the Archive who I can speak to about their experience reading its catalog’s metadata descriptions. To account for intragroup differences among visitors, we will seek out a selection of visitors with as diverse of identity characteristics as possible.}\newline\newline
    What:
    Ongoing work includes conducting historical research to understand the context in which the metadata descriptions were written. For example, employees at the Archive stated that for many years, people wrote metadata descriptions with the aim of being as neutral and objective as possible, however the latest generation of archivists is challenging this, arguing that neutrality isn't possible and encouraging transparency instead.
    
    \subsubsection*{Who or what is excluded from the research?}
    
    Who:
    \begin{itemize}
      \item Past employees of the Archive
      \item People represented in the Archive’s cultural heritage records
      \item The majority of the Archive’s visitors (the research only has the capacity to include a selection of visitors in user research and participatory action research activities)
    \end{itemize}
    What:
    The historical context of metadata descriptions written before my lifetime
    \newline\newline \textit{To Do: Determine if policy guidelines for the Archive since its beginnings in the 16\textsuperscript{th} century are available to understand how it perceived itself and what drove its collection and documentation practices. Otherwise, the historical existence of the Archive is also excluded form the research.}
    
    \subsubsection*{How will the research define knowledge?}
    
    The research will define knowledge as multifaceted. We (the research team) will draw on the disciplines of gender studies and linguistics to manually identify and annotate types of contextual gender bias in metadata descriptions. The research will share the annotated dataset as one interpretation of gender bias, recognizing that different people have different experiences of oppression that cause variations in attitudes towards words or phrases.
    
    We will use the annotated dataset to train a discriminative classification algorithm. The types of gender bias that the algorithm identifies will be presented as potentially biased text, requiring verification from a person working with the text to decide whether the text should be considered biased.
    
    \subsubsection*{Who has agency and who can be empowered?}
    
    We (the research team) have agency as the people applying NLP methods to the Archive’s metadata descriptions.
    
    The employees of the Archive can be empowered through participatory action research, with collaborative activities in which we situate the employees as partners in the research and as experts on archival practices and metadata.
    
    The employees of the Archive have determined that people who do not identify as male are underrepresented in the Archive’s collections and thus those collections’ metadata descriptions. We focus our bias identification and classification efforts on gender bias to explore how we can empower people who do not identify as male through the process and outputs of our NLP research.\newline\newline
    \textit{To Do: Provide examples of how our research process and outputs empowers people who do not identify as male.}

\end{appendices}

\bibliographystyle{coling}
\bibliography{situated}

\end{document}